\pdfoutput=1
\documentclass[11pt]{article}

\usepackage[]{EMNLP2022}

\usepackage{times}
\usepackage{latexsym}

\usepackage[T1]{fontenc}
\usepackage[utf8]{inputenc}

\usepackage{inconsolata}
\usepackage{times}
\usepackage{color}
\usepackage{latexsym}

\usepackage{graphicx}
\usepackage{multirow}
\usepackage{tabularx}
\usepackage{booktabs}
\usepackage{xspace}
\usepackage{amsmath,amsfonts,amssymb}
\usepackage{float}
\usepackage{slashed}
\usepackage{amsbsy}
\usepackage{amstext}
\usepackage{microtype}
\usepackage{pgfplots}
\pgfplotsset{width=1.0\columnwidth}
\usepackage[multiple]{footmisc}
\usepackage{enumitem}
\usepackage{tabularx}
\usepackage{bbding}

\providecommand{\tabularnewline}{\\}
\floatstyle{ruled}
\newfloat{algorithm}{tbp}{loa}
\providecommand{\algorithmname}{Algorithm}
\floatname{algorithm}{\protect\algorithmname}

\newcommand{\IID}{I.I.D.\xspace}
\newcommand{\QGG}{\textsc{QGG}\xspace}

\newcommand{\Transduction}{\textsc{Transduction}\xspace}
\newcommand{\Ranking}{\textsc{Ranking}\xspace}

\title{TIARA: Multi-grained Retrieval for Robust Question Answering over Large Knowledge Bases}
\author{Yiheng Shu$^{\heartsuit*}$, Zhiwei Yu$^\diamondsuit$, Yuhan Li$^{\clubsuit*}$, Börje F. Karlsson$^\diamondsuit$, Tingting Ma$^{\spadesuit*}$, \\
\textbf{Yuzhong Qu}$^\heartsuit$ \and \textbf{Chin-Yew Lin}$^\diamondsuit$ \\
$^\heartsuit$ State Key Laboratory of Novel Software Technology, Nanjing University, China; \\
$^\diamondsuit$ Microsoft Research; $^\clubsuit$ Nankai University; $^\spadesuit$ Harbin Institute of Technology\\
\texttt{yhshu@smail.nju.edu.cn}, \texttt{yzqu@nju.edu.cn}, \texttt{yuhanli@mail.nankai.edu.cn} \\
\texttt{\{zhiwei.yu,borje.karlsson,cyl\}@microsoft.com}
}

\begin{document}
\maketitle

\begin{abstract}

%Pre-trained language models (PLMs) have shown strong results across multiple NLP scenarios. However, KBQA remains a challenge, especially regarding coverage and generalization settings.
Pre-trained language models (PLMs) have shown their effectiveness %and generalization capability
in multiple scenarios. 
However, KBQA remains challenging, especially regarding coverage and generalization settings.
This is due to two main factors:
i) understanding the semantics of both questions and relevant knowledge from the KB;
ii) generating executable logical forms with both semantic and syntactic correctness.
In this paper, we present a new KBQA model, \textbf{TIARA}, which addresses those issues by
applying multi-grained retrieval to help the PLM focus on the most relevant KB contexts, viz., entities, exemplary logical forms, and schema items.
Moreover, constrained decoding is used to control the output space and reduce generation errors. 
Experiments over important benchmarks demonstrate the effectiveness of our approach. 
\textbf{TIARA} outperforms previous SOTA, including those using PLMs or oracle entity annotations, by at least 4.1 and 1.1 F1 points on GrailQA and WebQuestionsSP, respectively. %including PLM or oracle entity based
Specifically on GrailQA, \textbf{TIARA} outperforms previous models in all categories, with an improvement of 4.7 F1 points in zero-shot generalization.\footnote{Code is available at \url{https://github.com/microsoft/KC/tree/main/papers/TIARA}.}
\let\thefootnote\relax\footnotetext{*Work performed during their internships at Microsoft Research Asia.}

%Transferable generation capability from pre-trained language models sheds a light on question answering in challenging generalization settings.
%There are two key points to build a robust and effective model:
%1) understand the question and find semantically relevant knowledge from the KB, and
%2) generate executable logical forms with both semantic and syntactic correctness.
%In this paper, we present a new KBQA model \textbf{TIARA}.
%1) We use multi-grained retrieval to make the PLM focuses on the most relevant KB contexts including entities, exemplary logical forms, and schema items.
%2) We use constrained decoding to control the output space and reduce generation errors.
%Experiments demonstrate that TIARA outperforms previous SOTA by $4.1$ and $1.1$ points in F1 on the GrailQA and WebQuestionsSP datasets, respectively.
%It achieves SOTA on GrailQA in both i.i.d. and challenging generalization settings.\footnote{Code and data will be released upon paper acceptance.}

\end{abstract}

\section{Introduction}
\label{sec:introduction}

Knowledge base question answering (KBQA) has established itself as an important and promising research area as it greatly helps the accessibility and usability of existing large-scale knowledge bases (KBs). Such KBs contain abundant facts in a structured form, which can not only be accessed, but also reasoned over. KBQA bypasses the need for users to learn complex and burdensome formal query languages, and it allows users to leverage knowledge using natural language. 
% However, querying the KB requires a formal query, and people cannot directly ask questions and obtain expected answers from the KB. 
%However, querying the KB requires a formal query, which is more burdensome compared to using natural language.
%Question Answering over Knowledge Base (KBQA) provides a user-friendly way to query KB facts.
% which has attracted huge interest from both industry and academia (cite...)
%to solve the problem and provides a user-friendly way for users to query KB facts. 
While recent KBQA efforts have achieved interesting results, most previous studies target or assume an underlying strong correspondence between the distributions of schema items in test questions and training data, i.e., the i.i.d. assumption. However, this assumption does not hold, especially for large-scale KBs, which typically contain very large numbers of entities and schema items (classes and relations).\footnote{Our experiments involve more than 45M entities, 2K classes, and 6K relations. RDF Schema contains rdfs:Class (class) and rdf:Property (relation).} 
This extensive space is a critical challenge as it both allows for a myriad of novel compositions of schema items \cite{keysers20measuring}, i.e., requires compositional generalization.
It also immensely increases the likelihood of user queries, including previously unseen items or domains \cite{gu21beyond}, i.e., requires strong zero-shot generalization.
 
% The distribution of user questions is difficult to capture, and sampling training corpus from large spaces is data-inefficient.

%, are pre-trained on large-scale text corpus and can be fine-tuned to achieve impressive performance in different domains for different downstream tasks
% However, structured data is totally different from unstructured text and contains rich structural information. There is an inconsistency between the natural language in pre-training and the logical form of the downstream task. It leads to two challenges in using PLMs for semantic parsing tasks: Different from unstructured data in the pre-training phase, structured data contains rich structural information, which leads to two challenges in using PLMs for semantic parsing tasks.

%Which makes leveraging PLMs one possible solution to the generalization challenges.

Meanwhile, pre-trained language models (PLMs), such as T5 \cite{raffel20exploring} and GPT-3 \cite{brown20language}, have demonstrated notable success in many natural language processing (NLP) scenarios \cite{karpas22mrkl}. 
Many times they even show strong generalization capabilities. 
Inspired by the progress of PLMs on unstructured text-to-text tasks, researchers have explored semantic parsing of natural language to logical form utilizing PLMs \cite{poesia22synchromesh,xie22unified} to tackle the generalization challenges and improve system robustness. However, differently from unstructured data in the typical PLM pre-training phase, KBs represent rich semantics and complex structures, which lead to two challenges in using PLMs for KBQA:
i) \textbf{KB Grounding}: given a KB, how to understand the semantics of both the question and the relevant knowledge from the KB?
% link with several structured knowledge grounding tasks.
ii) \textbf{Logical Form Generation}: how to make sure syntax and semantics of generated logical forms conform to the KB specification and are executable (i.e., guarantee correctness)?
% how to ensure that the syntax and semantics conform to the specifications of the KB.

\begin{figure*}[t]
\centering
\includegraphics[width=0.9\linewidth]{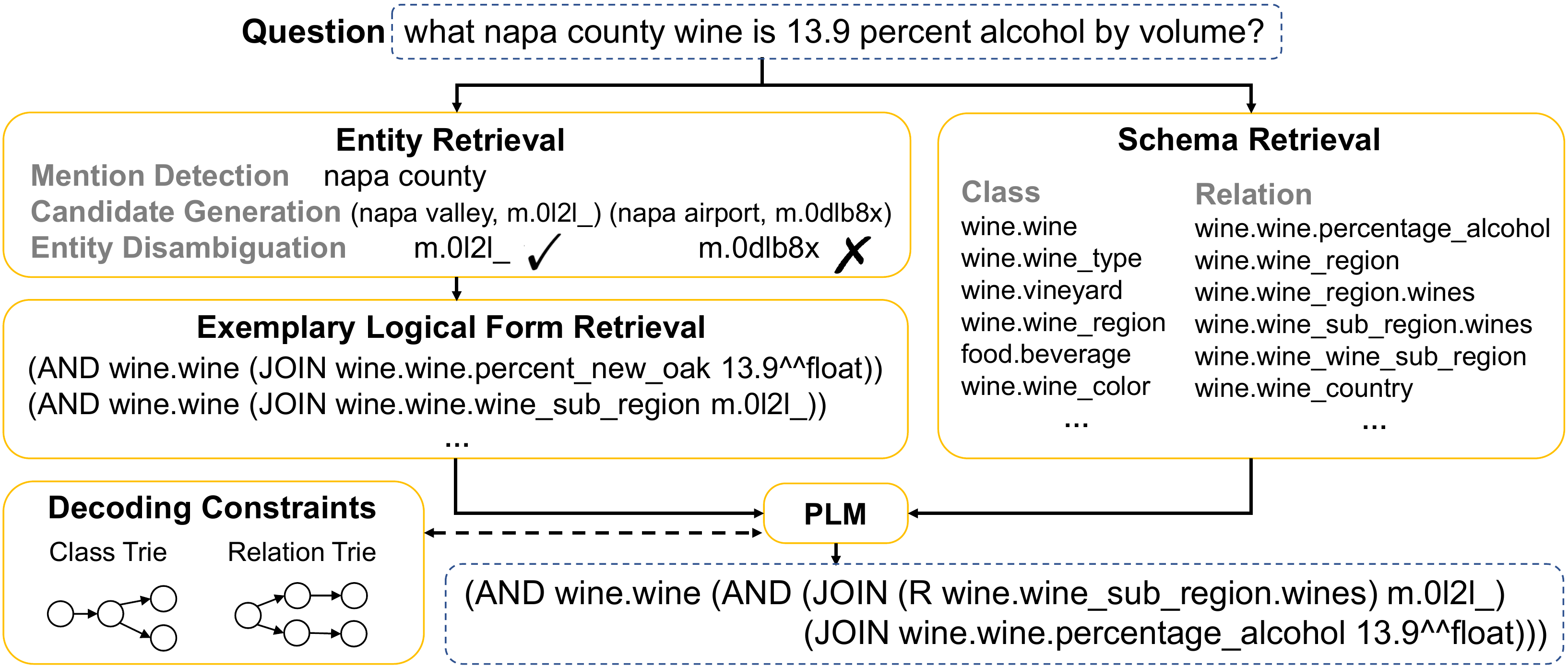}
\caption{
Overview of TIARA. 
% 1) Multi-grained retrieval, including entities, exemplary logical forms and schema items.
% 2) Constrained decoding controls the schema search space during target logical form generation.
1) \textit{Entity retrieval} grounds the mention to entity m.0l2l\_.
2) \textit{Exemplary logical form retrieval} enumerates logical forms starting from the entity m.0l2l\_ or the number 13.9, and ranks them.
3) \textit{Schema retrieval} independently grounds the most related schema items. % in the KB, including classes and relations.
4) Retrieved multi-grained contexts are then fed to the PLM for generation.
5) \textit{Constrained decoding} controls the schema search space during logical form generation.
}
\label{fig:tiara_overview}
\end{figure*}

KB grounding requires linking a question to relevant KB items, but the large size of KBs makes it challenging. 
We consider that contextual retrieval can help the PLM focus on the most relevant contexts \cite{xie22unified,wei22chain}.
Previous works propose different retrieval methods. 
\citet{das21case} and \citet{zan22s2ql} retrieve questions with logical form annotations that are similar to the input question from the training set, which is only effective for i.i.d. scenarios.
\citet{ye21rng} retrieve exemplary logical forms only within a two-hop range of linked entities.
\citet{chen21retrack} and \citet{xie22unified} retrieve schema items for each question without considering the connectivity on the KB and have poor zero-shot performance.
% , which cannot support the understanding of questions more than two hops or without topic entities. 

To make the best of structured KB contexts and be able to take advantage of recent PLM work,
we analyze all stages in a KBQA system to identify and address their challenges, and we demonstrate how leveraging multi-grained retrieval augments PLM generation - especially regarding robustness in compositional and zero-shot settings:
(\expandafter{\romannumeral1}) \textbf{Entity retrieval}: finding topic entities is a key step, and we augment the mention detection method to improve entity retrieval (linking) performance in zero-shot cases;
(\expandafter{\romannumeral2}) \textbf{Exemplary logical form retrieval}: logical forms provide the semantic and structural contexts from the KB, which assists the PLMs in KB grounding and in generating valid logical forms; and
(\expandafter{\romannumeral3}) \textbf{Schema retrieval}: logical form enumeration alone cannot properly support questions with more than two hops or diverse function types. However, retrieving schema items is not subject to this constraint and it can be used as a semantic supplement to logical forms.

% , including entity retrieval, structured exemplary logical form retrieval \cite{ye21rng}, and the semantic supplement from dense retrieval \cite{karpukhin20dense,chen21retrack}.
% including the syntax information stored in logical forms  
% Specifically, we leverage the entity linking paradigm \cite{}, the syntax information stored in logical forms \cite{ye21rng}, as well as the semantic supplement from dense retrieval \cite{karpukhin20dense,chen21retrack}. 

Although KB contexts are retrieved for the PLM, it may still generate invalid logical forms due to its unconstrained output space, e.g., generating schema items that do not exist in the KB. 
Similarly to how seq2seq models combined with rules \cite{liang17neural,chen21retrack} reduces syntax errors during logical form generation, we further introduce \textbf{constrained decoding} to alleviate this issue for the PLM using prefix trees constructed from the KB.

% \textbf{Constrained decoding} helps reduce generation errors that violate the rules of logical form, and we apply it with prefix trees constructed from the KB.

% In order to explore the capabilities of PLMs on the KB semantic parsing task, we synthesize the above KB context types and perform a pilot experiment.
% Our pilot experiment shows that given oracle entities, exemplary logical forms retrieved from oracles entities, and oracle schema items, T5-base \cite{raffel20exploring} can achieve up to $92.0\%$ F1 ($76.8\%$ for previous SOTA) on GrailQA \cite{gu21beyond} validation set, demonstrating PLM can take advantage of these contexts to achieve a strong generalization.
% Its details are discussed in Section \ref{subsec:future exploration}.

Here we propose a mul\textbf{TI}-gr\textbf{A}ined \textbf{R}etrieval \textbf{A}ugmented (\textbf{TIARA}) KBQA method (Section \ref{sec:method}) that addresses the two mentioned challenges.
%In this paper, we propose a mul\textbf{TI}-grained retrieval \textbf{A}ugmentation and const\textbf{RA}ined decoding (\textbf{TIARA}) method for KBQA.
As shown in Figure \ref{fig:tiara_overview}, we employ multi-grained retrieval to provide both semantic and syntactic references to a PLM.
% As shown in Figure \ref{fig:tiara_overview}, we leverage the \textbf{retrieved contexts} of three granularities to give both semantic and syntactic references for the PLM.
Then, target logical form generation is controlled by constrained decoding.
% TIARA further strengthens the potential of PLMs for KBQA, when compared to previous methods.
% We analyze the impact of providing PLMs with different granularities of contexts on different generalization levels and question types (Section \ref{sec:analysis}).
Utilizing these mechanisms, TIARA achieves improved KBQA performance not only in i.i.d., but also in compositional and zero-shot generalization (Section \ref{sec:experiments}).
Experiments over two important benchmarks demonstrate the effectiveness of our approach. TIARA outperforms all previous SOTA methods, including those using PLMs or even oracle entity annotations, by at least 4.1 and 1.1 F1 points on GrailQA and WebQuestionsSP, respectively.
Specifically on GrailQA, TIARA outperforms previous models in all generalization categories, with an improvement of 4.7 F1 points in zero-shot generalization.
To further explore how TIARA strengthens the potential of PLMs for KBQA, we also analyze the impact of providing PLMs with different granularities of contexts on different generalization levels and question types (Section \ref{sec:analysis}).

% The contributions are as follows.
% \begin{itemize}
%     \item We discuss the challenges of neural semantic parsing on large-scale KB and propose TIARA, which effectively improves PLM performance on KBQA.
%     \item We analyze the impact of providing PLMs with different granularities of contexts on different generalization levels and question types.
%     \item Experiments show that TIARA improves KBQA performance not only on i.i.d., but also in compositional and zero-shot generalization. 
%     TIARA outperforms previous SOTA by $4.1$ and $1.1$ points in F1 on the GrailQA and WebQuestionsSP datasets, respectively.
% \end{itemize}

% \begin{figure}[t]
% \centering
% \includegraphics[width=0.9\linewidth]{img/TIARA_MD.pdf}
% \caption{
% % An example of mention detection. 
% % Tokens are represented by BERT embeddings. 
% % The model learns the representation of each span and classifies whether it is an entity mention.
% SpanMD learns whether each possible span is an entity mention.}
% \label{fig:mention detection}
% \end{figure}

\section{Related Work}
\label{sec:related work}

\subsection{Knowledge Base Question Answering}
\label{subsec:knowledge base question answering}

KBQA endeavors to allow querying KBs through natural language questions.
Existing KBQA methods are mainly either information retrieval-based (IR-based) or semantic parsing-based (SP-based) methods \cite{lan21complex}.
IR-based methods \cite{shi21transfer,zhang22subgraph} construct a question-specific graph and rank entities or paths to get the top answers. % They do not require logical form annotations.
% , but their blackbox style reasoning is hard to explain.
SP-based methods aim at generating target logical forms, which makes them more interpretable. %  (e.g., query graph, SPARQL, s-expression)
Such methods can be classified into feature-based ranking and seq2seq generation methods.
Feature-based ranking methods search and rank query graphs in a step-wise manner.
\citet{yih15semantic} define a query graph and propose a staged generation method.
\citet{lan20query} incorporate constraints during the staged query graph generation. % and prune the search space.
\citet{hu21edg} use rules to decompose questions and guide the search.
% , but rules are hard to cover diverse questions.
% It is difficult for feature-based ranking methods to prune the search space when the number of hops grows.
% Seq2seq generation methods \cite{liang17neural,chen21retrack} use sequence-to-sequence models to map natural language questions to logical forms.
Seq2seq generation methods \cite{liang17neural,chen21retrack} convert natural language sequences to logical form sequences.
They are more flexible than the above ranking methods in generating logical forms with more hops or functions.
In particular, seq2seq generation methods using PLMs \cite{oguz2020unik,das21case,ye21rng,cao22kqa,hu2022logical} show promising performance.
Among them, the most relevant work to ours is RnG-KBQA \cite{ye21rng}, which also retrieves exemplary logical forms as important contexts.
However, TIARA further explores the use of PLMs by improving zero-shot mention detection, matching entity-independent semantics with schema retriever, and reducing generation errors with constrained decoding.
% and address their limitations to enhance the robustness of KBQA methods. %address the challenge of generalization.

\subsection{Neural Semantic Parsing with PLMs}
\label{subsec:neural semantic parsing}

Several existing neural semantic parsing methods consider contextual retrieval when applying PLMs. % and constrained decoding when applying PLM.
\citet{shi22nearest} use nearest neighbor retrieval to augment zero-shot inference.
\citet{xie22unified} feed contexts to PLMs and conduct experiments on several semantic parsing tasks over structured data.
Furthermore, other works also leverage constrained decoding for semantic parsing.
\citet{picard21scholak} propose multiple levels of constrained decoding for the text-to-SQL task and improve performance without additional training.
\citet{richard21constrained} illustrate PLMs with constrained decoding have few-shot semantic parsing ability. 
While, \citet{poesia22synchromesh} propose that retrieval of related examples and constrained decoding can improve the performance of the code generation task without fine-tuning.
% In this paper, we introduce the above idea of using PLMs for semantic parsing on KB.
In this paper, we introduce the idea of multi-grained contextual retrieval along with constrained decoding to bolster KBQA scenarios.

\section{Method}
\label{sec:method}

% Our method starts with the multi-grained retrieval, as shown in Figure \ref{fig:tiara_overview}.
% , followed by target logical form generation, 
% Entities and schema items are the basic elements in a KB. 
% Multi-grained retrieval starts from these two types of elements respectively.
To demonstrate how comprehensive KB context enhances the robustness of retrieval, a key component of our method, we retrieve KB contexts in a multi-grained manner and then generate target logical forms with constrained decoding, as shown in Figure \ref{fig:tiara_overview}.
We introduce multi-grained retrieval in Section \ref{subsec:multi-grained retriever}, target logical form generation in Section \ref{subsec:logical form generation}, and describe constrained decoding in Section \ref{subsec:constrained decoding}.
% Entity retrieval finds entity IDs and labels, and logical form retrieval finds exemplars within the entity's neighborhood.
% % , including enumeration and ranking.
% % These structured logical forms provide exemplars for the target logical form generation.
% In addition, schema retrieval finds schema items that highly match the semantics of the question.
% % and assists PLMs in question understanding. 
% Then, these retrieved contexts (Section \ref{subsec:multi-grained retriever}) are provided to PLMs to generate the target logical form (Section \ref{subsec:logical form generation}).
% During the PLM decoding, constrained decoding is applied to reduce invalid generation (Section \ref{subsec:constrained decoding}).

\subsection{Preliminaries}
\label{subsec:preliminaries}
Here, a knowledge base is an RDF graph consisting of a collection of triples in the form ($s$, $r$, $o$), where $s$ is an entity, $r$ is a relation, and $o$ can be an entity, a literal, or a class.
We use s-expressions as the logical form following \citet{gu21beyond} and \citet{ye21rng}, which can later be converted to SPARQL queries and executed over KBs.
In this paper, \textbf{exemplary logical form} refers to the logical form that is input to the generative PLM as the context. It is obtained by enumeration and ranking on KBs.
\textbf{Schema} refers to rdfs:Class (class) and rdf:Property (relation) together, and they are the necessary elements of logical forms.
\textbf{Target logical form} refers to the logical form generated by the PLM.
Examples of above concepts are shown in Figure \ref{fig:tiara_overview}.

\subsection{Multi-grained Retriever}
\label{subsec:multi-grained retriever}

% In the subsection, we present how linked entities and top-ranked logical forms and schema are retrieved for the following generation (Section \ref{subsec:logical form generation}).

\paragraph{Entity Retriever}
% The ability to retrieve relevant KB entities referred to in a question is fundamental for the task of KBQA. 
%
We perform entity retrieval following a standard pipeline with three steps, i.e., mention detection, candidate generation, and entity disambiguation \cite{shen2021entity}. To better detect zero-shot mentions, we regard mention detection as a span classification task \cite{boningknife}. % , as shown in Figure \ref{fig:mention detection}.
In this way, we can deal well with sentences containing out-of-vocabulary entities \cite{fu20spanner} and achieve higher recall by adjusting the threshold of outputting a candidate mention \cite{yu20neural}. 
%SpanNER can deal with sentences containing out of vocabulary words well \cite{fu20spanner} and achieve high recall by adjusting the threshold \cite{yu20neural}. %Inspired by \citet{shen2021entity}, 
%To better detect zero-shot mentions, we regard the mention detection as a span classification task \cite{shen2021entity}, as shown in Figure \ref{fig:mention detection}. 
We obtain token representations from contextualized pre-trained BERT \cite{devlin19bert} and 
%Then, bidirectional LSTM \cite{hochreiter1997long} is applied to learn the token representations. 
enumerate all the possible spans in the question within a maximum length. 
For each span, representations of the start and the end token are concatenated together with the length embedding as the span representation.\footnote{Length embedding is obtained by a learnable look-up table.} 
Span representations are fed into a classifier (i.e., softmax layer) to get the probability that each span is a mention. 
We refer to our mention detector as SpanMD in the rest of the paper.
%In the mention detection step, we employ the SpanNER \cite{shen2021entity}, a SOTA Named Entity Recognition (NER) model, to detect all possible text spans that can be linked to entities. %As Fig XX shows, we obtain the token embedding by feeding the word to 
%
%We first enumerate all the possible spans for the given question and then re-assign a label for each span to indicate whether it is detected as a mention.
%
%It is worth noting that different from the BERT-NER\footnote{ \url{https://github.com/kamalkraj/BERT-NER}.} which conceptualizes Named Entity Recognition (NER) as a sequence labeling task and is widely used by previous KBQA works \cite{gu21beyond,chen21retrack,ye21rng}, we employ SpanNER \cite{yu20named} for mention detection. 
%SpanNER shows its superiority in dealing with sentences containing more out-of-vocabulary (OOV) words \cite{fu20spanner} %and can obtain a higher recall by adjusting the threshold, 
%which is helpful in tackling zero-shot cases.%the following two scenarios.
%
%Firstly, SpanNER is better at dealing with entities locating on sentences with more Out-of-Vocabulary (OOV) words \cite{fu21spanner}, which is helpful in tackling zero-shot cases.
%
%Secondly, SpanNER can obtain a higher recall via controlling a suitable length threshold of the span. This allows us to detect more mentions and thus better understand the semantics of the question. 
%As for the candidate generation step,
%Given detected mentions, 
Then, we generate candidate entities for each mention using the alias mapping FACC1 \cite{gabrilovich13facc1}.
For entity disambiguation, we leverage a cross-encoder framework as \citet{ye21rng} to jointly encode the question and contexts of each candidate entity (i.e., the entity label and its linked relations) and obtain a matching score. 
The candidate with the highest score is kept as the entity for each mention. %we adopt the model proposed by \citet{ye21rng}. To be specific, we leverage a cross-encoder framework to jointly encode the question and the contexts of each candidate entity (i.e., the label of the entity together with its relations). We obtain the score in the same way as described in the next section. The candidate with the highest score is kept as the entity for the mention

\begin{figure}[t]
\centering
\includegraphics[width=\linewidth]{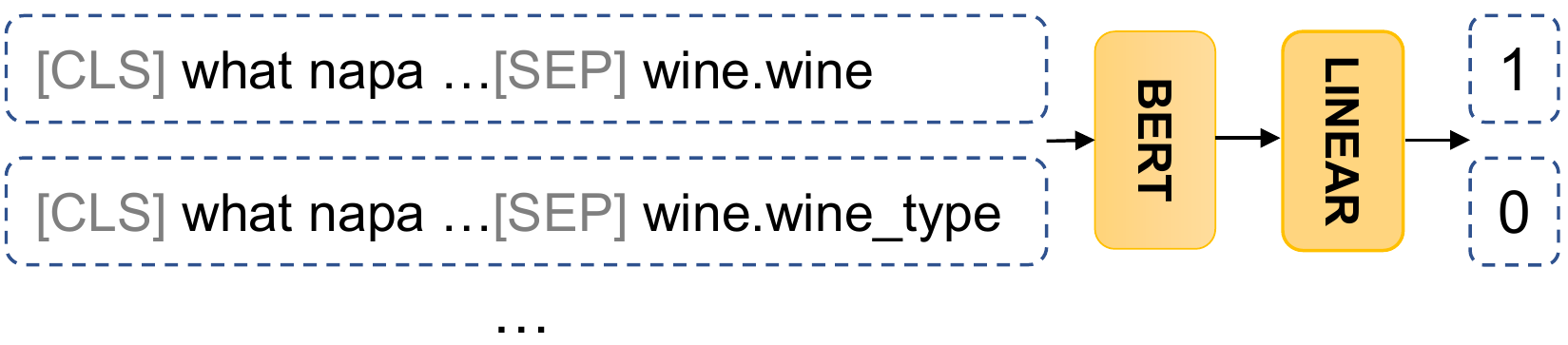}
\caption{
% A running example of schema retrieval. 
% The question and each schema is concatenated by \texttt{[SEP]} token. 
% The \texttt{[CLS]} output of BERT is fed to a linear layer and used for classification.
Schema retrieval learns if a question and a schema item are a match or not.
}
\label{fig:schema retrieval}
\end{figure}

\paragraph{Exemplary Logical Form Retriever}
The logical form retrieval includes enumeration and ranking.
% We follow the RnG-KBQA \cite{ye21rng} where the logical form retrieval consists of enumeration and ranking.
We follow the enumeration method proposed by \citet{gu21beyond} and \citet{ye21rng}.
Starting from all potential entities, it searches their neighborhood up to two hops since it is too costly to enumerate beyond that due to the exponential increase in candidates.
%\footnote{As the candidate number increases exponentially, it would take one month to train 3 hops under the same settings.}.
Each path is converted to an s-expression as an exemplary logical form.
For ranking, each pair $(x, c)$ of question $x$ and candidate exemplary logical form $c$ is concatenated by a \texttt{[SEP]} token and used as input to BERT.
The \texttt{[CLS]} representation is fed into a linear layer to get the score $s(x,c)$ as follows:

\begin{equation}
    s(x, c) = \textsc{Linear}(\textsc{BertCls}([x;c])) 
    \label{equ:bert_linear}
\end{equation}

To distinguish the target logical form from the enumeration results, the ranker is optimized to minimize the following loss function:

\begin{equation}
    \mathcal{L}_{\text{ranker}}=-\frac{e^{s(x, y)}}{e^{s(x, y)}+\sum_{c \in C \wedge c \neq y} e^{s(x, c)}}  
\end{equation}
where $y$ is the target logical form or its semantic equivalent.
Note that a question may not contain any entity, in which case no exemplary logical forms will be retrieved.
Due to the hop limitation and the diversity of functions, the enumeration is not guaranteed to cover target logical forms.

\paragraph{Schema Retriever}
% The exemplary logical form retrieval is not applicable for specific cases.
Independent of the above entity and exemplary logical form retrieval, we employ dense schema retrieval to find semantically relevant items (classes and relations), which are not restricted to the neighbors of entities and can be used as a semantic supplement to exemplary logical forms. % compared to the exemplary logical form retrieval.
As shown in Figure \ref{fig:schema retrieval}, schema retrieval is implemented by a cross-encoder.
% with the sentence pair classification task.
Compared to the bi-encoder \cite{chen21retrack}, it learns the interaction representation between the question and schema items.
The matching score $s(x, c)$ of a question $x$ and a schema $c$ is calculated as Equation \ref{equ:bert_linear}.
% We concatenate a question $x$ and a schema $s$ by \texttt{[SEP]} token and input it to BERT.
% The output of the \texttt{[CLS]} token is fed into a linear layer to classify whether the question and the schema match as follows:
The objective is the same as in the sentence-pair classification task \cite{devlin19bert}.
We select top-ranked classes and relations with the highest scores for each question.
Class and relation retrievers are trained separately.
% Negative samples for each question are randomly sampled.

\begin{table*}[t]
\small
\centering
\begin{tabular}{lcccccccc}
\toprule
&\multicolumn{2}{c}{\textbf{Overall}}   &\multicolumn{2}{c}{\textbf{\IID}}  &\multicolumn{2}{c}{\textbf{Compositional}}    &\multicolumn{2}{c}{\textbf{Zero-shot}}            \\
 \cmidrule(lr){2-3} \cmidrule(lr){4-5} \cmidrule(lr){6-7} \cmidrule(lr){8-9}
\multicolumn{1}{l}{\textbf{Method}} & \textbf{EM} & \textbf{F1} & \textbf{EM} & \textbf{F1} & \textbf{EM} & \textbf{F1} & \textbf{EM} & \textbf{F1} \\ 
\midrule
\multicolumn{1}{l}{GloVe + \Transduction~\citep{gu21beyond}} & 17.6  & 18.4  & 50.5  & 51.6 & 16.4  & 18.5     & ~~3.0  & ~~3.1    \\
\multicolumn{1}{l}{\QGG~\citep{lan20query}} & - & 36.7  & - & 40.5  & -  & 33.0  & - & 36.6    \\
\multicolumn{1}{l}{BERT + \Transduction~\citep{gu21beyond}} & 33.3  & 36.8  & 51.8  & 53.9  & 31.0  & 36.0     & 25.7  & 29.3    \\
\multicolumn{1}{l}{GloVe + \Ranking~\citep{gu21beyond}}     & 39.5  & 45.1  & 62.2  & 67.3 & 40.0  &  47.8 & 28.9  & 33.8  \\
\multicolumn{1}{l}{BERT + \Ranking~\citep{gu21beyond}}      & 50.6  & 58.0 & 59.9  & 67.0 & 45.5  &  53.9  & 48.6  & 55.7  \\
\multicolumn{1}{l}{ReTraCk~\citep{chen21retrack}} & 58.1 & 65.3 & 84.4 & 87.5 & 61.5 & 70.9 & 44.6 & 52.5  \\
\multicolumn{1}{l}{S$^{2}$QL~\citep{zan22s2ql}}   & 57.5 & 66.2 & 65.1 & 72.9 & 54.7 & 64.7 & 55.1 & 63.6 \\
\multicolumn{1}{l}{ArcaneQA~\citep{gu22dynamic}}  & 63.8 & 73.7 & 85.6 & 88.9 & 65.8 & 75.3 & 52.9 &  66.0 \\
\multicolumn{1}{l}{RnG-KBQA~\citep{ye21rng}}      & 68.8 & 74.4 & 86.2 & 89.0 & 63.8 & 71.2 & 63.0 & 69.2 \\
\midrule

\multicolumn{1}{l}{\textbf{TIARA} (Ours)} & \textbf{73.0} & \textbf{78.5} & \textbf{87.8} & \textbf{90.6} & \textbf{69.2} & \textbf{76.5} & \textbf{68.0} & \textbf{73.9} \\
% \multicolumn{1}{l}{\textbf{TIARA} } & $72.1$ & $77.5$ & $87.2$ & $90.2$ & $68.9$ & $76.1$ & $66.6$ & $72.3$ \\
\bottomrule
\end{tabular}
\caption{EM and F1 results (\%) on the hidden test set of GrailQA. 
% The results of other methods are taken from the leaderboard. 
TIARA outperforms other methods with three levels of generalization settings in both EM and F1.
}
\label{table:grailqa test results}
\end{table*}

\begin{figure}[t]
\centering
\includegraphics[width=\linewidth]{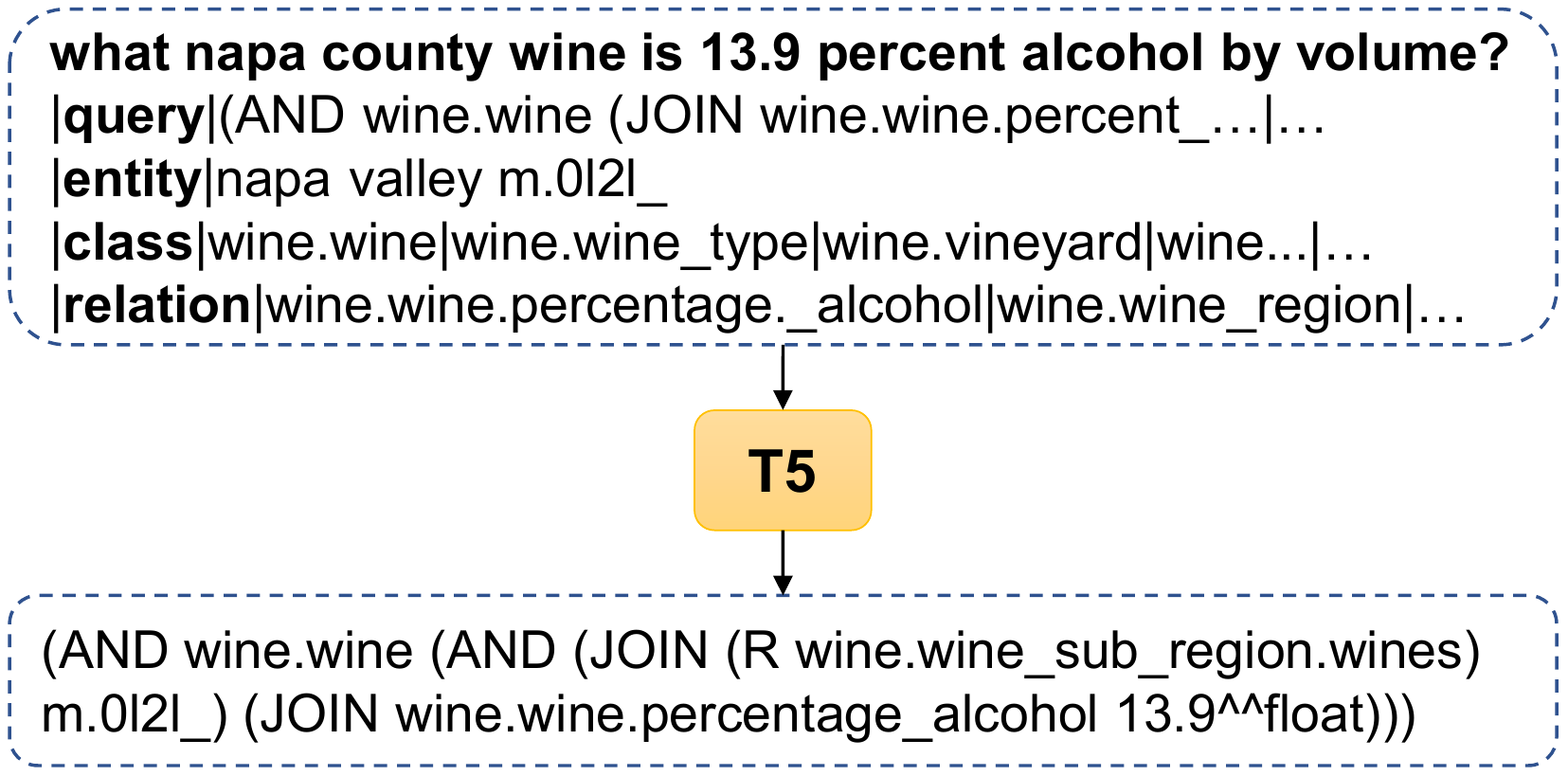}
\caption{
% A running example of target logical form generation.
% The retrieved contexts are concatenated and fed to T5. 
Given a set of retrieved contexts, T5 generates the target logical form.
}
\label{fig:target logical form generation}
\end{figure}

\subsection{Target Logical Form Generation}
\label{subsec:logical form generation}

The previously retrieved contexts are used to augment the PLM in target logical form generation. Generation is performed by a transformer-based seq2seq model - T5 \cite{raffel20exploring}.
As shown in Figure \ref{fig:target logical form generation},
the input sequence is a concatenation of the input question, retrieved entities, exemplary logical forms, and schema items.
The entity is represented by its label and ID.
The output sequence is the target logical form.
% \begin{equation}
% \begin{aligned}
%      q |\text{query}|l_1| \cdots |l_k| \text{entity} |e_1| \cdots |e_j| \\
%      \text{class}|c_1| \cdots |c_m| \text{relation} |r_1| \cdots |r_n
% \end{aligned}
% \end{equation}
% where $q$ is the question, $l_i$ is a candidate logical form, $e_i$ is label and ID of an entity, $c_i$ is a class and $r_i$ is a relation. 
% The labels \textit{query}, \textit{entity}, \textit{class}, and \textit{relation} are used to indicate the type of the following content.
The T5 model is fine-tuned to generate the target sequence with the cross-entropy objective as follows:

\begin{equation}
    \mathcal{L}_{\text{gen}}=-\sum_{t=1}^{n} \log \left(p\left(y_{t} | y_{<t}, x, c\right)\right)
\end{equation}
where $x$ denotes the input question, $c$ denotes retrieved contexts, $y$ denotes the target sequence, and $n$ is the length of the target sequence.
During inference, the model uses beam search and retains the top-$k$ sequences.
Constrained decoding is performed during beam search (Section \ref{subsec:constrained decoding}).

\subsection{Constrained Decoding}
\label{subsec:constrained decoding}

\begin{figure}
    \centering
    \includegraphics[width=\linewidth]{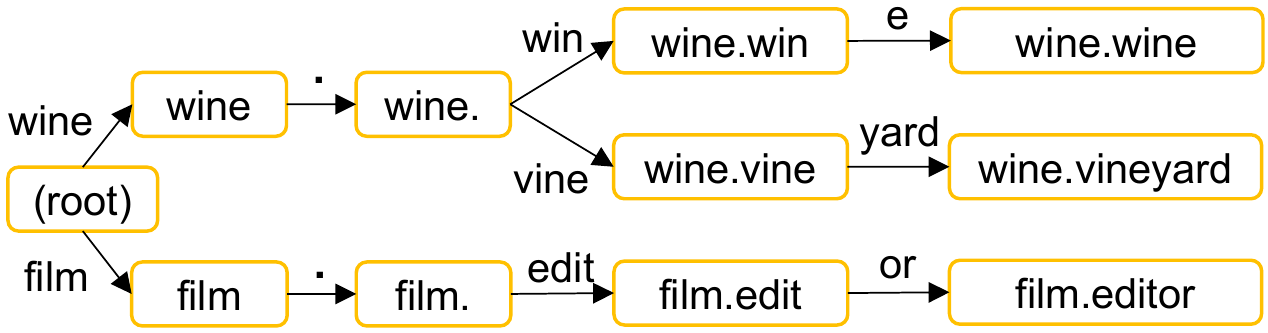}
    \caption{An example of a trie (prefix tree) that stores KB classes. 
    Each edge represents a token that the PLM can select.
    %The bolded nodes denote schema items.
    }
    \label{fig:trie}
\end{figure}

% The unconstrained output space over the entire vocabulary and the token-by-token decoding makes PLM possible to violate the logical formal syntax or generate items that do not exist in KB. Therefore,
% PLM decodes token-by-token with beam search.
Directly applying the beam search algorithm in the decoding process may generate invalid operators or schema items.
% Given a sequence of partial outputs, we assign the score $-\infty$ to invalid tokens to control the search space.
The constrained decoding focuses on reducing generation errors on logical form operators and schema items.
% The beam search ranks sequences by the average score of each token.
% During beam search, the decoding constraint is implemented by assigning the score $-\infty$ to invalid tokens.
% It excludes some invalid sequences during the search process rather than filtering invalid sequences after top-k sequences are generated.
% As PLM decodes token by token, constraints are implemented on the token level.
For operators, only those operator tokens allowed by the s-expression syntax are considered valid during the generation process.
For schema items, their tokens are stored in the trie (prefix tree), as shown in Figure \ref{fig:trie}.
KB classes and relations are stored in two separate tries.
When a constraint rule finds that a schema item is currently being decoded, only the options in the trie are valid for the next token.

% The real execution tries to convert the generated logical forms of beam search to SPARQL query one by one, executes them, and checks if there is an answer.
% In this case, we then check the retrieved logical forms one by one.
% The above checking is done until the first logical form is executable and has an answer.

\section{Experiments}
\label{sec:experiments}

\subsection{Setup}
\label{subsec:setup}

\paragraph{Datasets} We perform experiments on two important standard KBQA datasets. 

\noindent\textbf{GrailQA} \cite{gu21beyond} is a large-scale KBQA dataset on Freebase \cite{bollacker08freebase} with 64,331 questions annotated with logical forms. 
It is carefully split to evaluate three levels of generalization (i.i.d., compositional, and zero-shot).
Questions contain up to 4-hop relations and optionally contain functions for counting, superlatives (\texttt{ARGMIN}, \texttt{ARGMAX}), and comparatives ($<$, $\leq$, $>$, $\geq$).

\noindent\textbf{WebQuestionsSP} (WebQSP) \cite{yih16thevalue} is a widely used semantic parsing dataset containing 4,937 questions from Google Suggest API.
% KBQA dataset on Freebase in i.i.d. setting with 4,937 questions.
We follow the split of training and validation sets as \citet{ye21rng} since there is no official split.

\paragraph{Evaluation Metrics}
% For both datasets, we choose s-expression \cite{gu21beyond} as the generation target 
We use the official scripts to evaluate on both datasets.
For GrailQA, we use exact match (EM) and F1 score (F1) as metrics. For WebQSP, we report F1 score and hits@1. It is worth noting that SP-based methods, including TIARA, generate logical forms that yield unordered answers, which are usually evaluated by F1, not hits@1. For a comprehensive comparison, we randomly select an answer for each question 100 times and calculate the average hits@1.\footnote{Previous SP-based methods calculate hits@1 by random selection with unknown times, so results may fluctuate.}
%For a comprehensive comparison of IR-based and SP-based methods on WebQSP, we calculate hits@1 by selecting a random answer 100 times for each question and calculate the average number of times it appears in the correct answer
% For comparison, the hits@1 metric of TIARA on WebQSP is obtained by randomly selecting one predicted answer 100 times for each question, because answers obtained by executing logical forms rather than ranking entities are unordered.

\paragraph{Implementation Details}
Our experiments are done on a machine with an NVIDIA A100 GPU and 216GB of RAM.
We implement our models utilizing PyTorch \cite{paszke19pytorch} and Hugging Face.\footnote{\url{https://huggingface.co/}}
Due to the significant difference in dataset sizes, their experimental setups are different.
For entity retriever, we use our proposed retriever (Section \ref{subsec:multi-grained retriever}) for GrailQA with a maximum mention length of 15, and the off-the-shelf entity linker ELQ \cite{li20efficient} for WebQSP.\footnote{Using ELQ is analogous to previous SOTA. The domain and mention length distribution between the two datasets is significant.}
For schema retriever, we fine-tune the BERT-base-uncased model for at most 3 epochs with early stopping. 
We randomly sample 50 negative candidates for each question. 
The top-10 classes and relations of the highest scores are kept respectively. 
As generator, we use the T5-base seq2seq generation model and fine-tune it with a learning rate of 3e-5. 
The beam size is set to 10. 
We limit the maximum number of tokens to 1,000 for inputs and 128 for outputs. 
For GrailQA, we train the generation model for 10 epochs with a batch size of 8. 
For WebQSP, we train it for 20 epochs with a batch size of 2.
After generation, we first check the logical forms generated by T5 in the beam search order by execution and exclude non-executable or invalid queries (with null answers).
Then, if no generated logical form is valid, we further check if any exemplary logical form is valid.
The first valid query is taken as the final prediction. % https://arxiv.org/abs/1807.03100

\begin{table}[t]
\centering %
\small
\begin{tabular}{lc@{\hspace{3pt}}c}
\toprule 
\textbf{Method} & \textbf{F1} &  \textbf{Hits@1}\tabularnewline
\midrule 
\multicolumn{3}{c}{\textit{IR-based methods}}\tabularnewline
EmbedKGQA$^*$~\citep{saxena20improving} & - & 66.6 \tabularnewline
GRAFT-Net~\citep{sun18open} & 62.8 & 67.8 \tabularnewline
PullNet~\citep{sun19pullnet} & - & 68.1 \tabularnewline
TransferNet${}^\heartsuit$~\citep{shi21transfer} & - & 71.4 \tabularnewline
% EmbedKGQA$^*{}^\heartsuit$~\citep{saxena20improving} & - & 72.5 \tabularnewline
Relation Learning${}^\heartsuit{}^\clubsuit$~\citep{yan21large} & 64.5 & 72.9 \tabularnewline
NSM$^*{}^\heartsuit$~\citep{he21improving} & 67.4 & 74.3 \tabularnewline
Subgraph Retrieval$^*$~\citep{zhang22subgraph} & 74.5 & \textbf{83.2} \tabularnewline
\midrule
\multicolumn{3}{c}{\textit{SP-based (feature-based ranking) methods}}\tabularnewline
TextRay$^\heartsuit$~\citep{bhutani19learning} & 60.3 & - \tabularnewline
Topic Units$^\heartsuit$~\citep{lan19knowledge} & 67.9 & - \tabularnewline
UHop~\citep{chen19uhop} & 68.5 & -\tabularnewline
GrailQA \Ranking\!\!$^*{}^\heartsuit{}^\clubsuit$ ~\citep{gu21beyond} & 70.0 & - \tabularnewline
STAGG$^\heartsuit$~\citep{yih16thevalue} & 71.7 & -\tabularnewline
QGG$^\heartsuit$ \citep{lan20query} & 74.0 & -\tabularnewline
\midrule
\multicolumn{3}{c}{\textit{SP-based (seq2seq generation) methods}}\tabularnewline
NSM$^\heartsuit$~\citep{liang17neural} & 69.0 & -
\tabularnewline
ReTraCk~\citep{chen21retrack} &  71.0 & 71.6
\tabularnewline
CBR-KBQA~\citep{das21case} & 72.8 & - \tabularnewline
ArcaneQA~\citep{gu22dynamic} & 75.6 & - \tabularnewline
RnG-KBQA~\citep{ye21rng} &  75.6 & -
\tabularnewline
Program Transfer$^*{}^\clubsuit$~\citep{cao22program} & 76.5 & 74.6 \tabularnewline
\midrule 
% \multicolumn{2}{c}{\textit{Oracle Linking}}\tabularnewline
% Ours$^*$ &  & \tabularnewline
% ~~~~-- Schema  &  &  \tabularnewline
% ~~~~-- Checker &  &  \tabularnewline
% \midrule
% \multicolumn{2}{c}{\textit{No Oracle Linking}}\tabularnewline
\textbf{TIARA} (Ours) & \textbf{76.7} & 73.9  \tabularnewline
~~~~w/o Schema & 76.4 & 73.7 \tabularnewline
~~~~w/o ELF & 75.0 & 73.4 \tabularnewline
~~~~w/o ELF \& Schema & 73.2 & 71.1 \tabularnewline
\textbf{TIARA}$^*$ & \textbf{78.9} & 75.2  \tabularnewline
~~~~w/o Schema & 78.8 & 75.0 \tabularnewline
~~~~w/o ELF & 76.2 & 74.5 \tabularnewline
~~~~w/o ELF \& Schema & 75.4 & 73.1 \tabularnewline
\bottomrule 
\end{tabular}
\caption{F1 and hits@1 results (\%) on WebQSP. 
$*$ denotes using oracle entity linking annotations. 
$\heartsuit$ denotes the assumption of a fixed number of hops. 
$\clubsuit$ denotes pre-training on an auxiliary task or other KBQA datasets.
% NSM \cite{he21improving} denotes Neural State Machine, and another NSM \cite{liang17neural} denotes Neural Symbolic Machines.
For comparison, hits@1 on TIARA is obtained by randomly selecting one answer for each question 100 times.}
\label{table:webqsp test results}
\end{table}

\paragraph{Baselines for Comparison}
We compare against models on the GrailQA leaderboard and previous SOTA methods for WebQSP.\footnote{\url{https://dki-lab.github.io/GrailQA/} on Jun. 3, 2022.}
% For WebQSP, we compare with previous SOTA methods, including both IR-based and SP-based methods.
The results are taken from their published papers.
% Among these methods, ArcaneQA \cite{gu22dynamic} and RnG-KBQA \cite{ye21rng} focus on generalization challenges, and they are the methods we focus on for comparison.
RnG-KBQA is the previous SOTA on both GrailQA and WebQSP and uses a T5-base generator, the same as TIARA.

\begin{table*}[th]
\small
\centering

\begin{tabular}{lcccccccc}
\toprule
& \multicolumn{2}{c}{\textbf{Overall}}   &\multicolumn{2}{c}{\textbf{\IID}}  &\multicolumn{2}{c}{\textbf{Compositional}}    &\multicolumn{2}{c}{\textbf{Zero-shot}}            \\
 \cmidrule(lr){2-3} \cmidrule(lr){4-5} \cmidrule(lr){6-7} \cmidrule(lr){8-9}
\multicolumn{1}{l}{\textbf{Method}} & \textbf{EM} & \textbf{F1} & \textbf{EM} & \textbf{F1} & \textbf{EM} & \textbf{F1} & \textbf{EM} & \textbf{F1} \\ 
\midrule
% \multicolumn{1}{l}{BERT + \Ranking~\citep{gu21beyond}}  & 51.0 & 58.4 & 58.6 & 66.1 & 40.9 & 48.1 & 51.8 & 59.2  \\
\multicolumn{1}{l}{RnG-KBQA~\cite{ye21rng}}  & 71.4 & 76.8 & 86.7 & 89.0 & 61.7 & 68.9 & 68.8 & 74.7  \\
\multicolumn{1}{l}{ArcaneQA~\cite{gu22dynamic}}  & 69.5 & 76.9 & 86.1 & 89.2 & 65.5 & 73.9 & 64.0 & 72.8  \\ \midrule
\multicolumn{1}{l}{\textbf{TIARA} (Ours)} & \textbf{75.3} & \textbf{81.9} & \textbf{88.4} & \textbf{91.2} & \textbf{66.4} & \textbf{74.8} & \textbf{73.3} & \textbf{80.7} \\
~~~~w/o CD & 75.1 & 81.5 & 88.4 & 91.2 & 66.4 & 74.6 & 72.9 & 80.2  \tabularnewline
~~~~w/o Schema & 73.5 & 79.2 & 86.3 & 89.9 & 64.7 & 72.7 & 71.6 & 77.3 \tabularnewline
~~~~w/o Schema \& CD & 73.3 & 79.0 & 86.1 & 89.6 & 64.1 & 72.1 & 71.5 & 77.2 \tabularnewline
~~~~w/o ELF  & 62.3 & 65.0 & 87.7 & 88.9 & 60.4 & 66.0 & 51.9 & 54.2  \tabularnewline
~~~~w/o ELF \& CD & 60.2 & 62.6 & 87.3 & 88.6 & 59.2 & 64.4 & 48.8 & 50.6 \tabularnewline
~~~~w/o ELF \& Schema & 37.3 & 39.8 & 86.0 & 88.1 & 58.6 & 66.3 & \ \ 7.3 & \ \ 7.8  \tabularnewline
~~~~w/o ELF \& Schema \& CD & 32.9 & 34.6 & 85.1 & 86.7 & 52.2 & 57.7 & \ \ 2.1 & \ \ 2.3 \tabularnewline
~~~~w/o SpanMD w/ BERT-NER & 73.3 & 79.7 & 87.4 & 90.0 & 66.3 & 74.7 & 70.0 & 77.3\tabularnewline
Exemplary Logical Forms & 67.2 & 72.9 & 72.8 & 76.7 & 55.3 & 60.7 & 69.7 & 76.3 \tabularnewline 

% T5-large {'em': 0.7691852728079255, 'f1': 0.83011552736624, 'em_iid': 0.8913998744507219, 'f1_iid': 0.9187115658424029, 'em_comp': 0.6928665785997358, 'f1_comp': 0.7678080794940922, 'em_zero': 0.7475382932166302, 'f1_zero': 0.8173146484783571}

% before mention detection
% em': 0.7328108827443442, 'f1': 0.7969578526616057, 'em_iid': 0.8738229755178908, 'f1_iid': 0.9004452976395948, 'em_comp': 0.6631439894319683, 'f1_comp': 0.7465517037365419, 'em_zero': 0.700218818380744, 'f1_zero': 0.7727399668909849}
% ~~~~-- DC & $73.3$ & $79.5$ & $87.3$ & $89.9$ & $66.4$ & $74.4$ & $70.0$ & $77.1$ \tabularnewline
% ~~~~-- Schema & $72.6$ & $78.0$ & $\mathbf{87.4}$ & $89.9$ & $64.9$ & $72.8$ & $69.4$ & $75.0$  \tabularnewline
% ~~~~-- Schema \& DC & $72.4$ & $77.8$ & $87.3$ & $89.7$ & $64.6$ & $72.3$ & $69.2$ & $74.8$  \tabularnewline
% ~~~~-- LF  & $60.5$ & $63.4$ & $85.8$ & $87.2$ & $58.5$ & $64.0$ & $50.4$ & $52.8$  \tabularnewline
% ~~~~-- LF \& DC & $58.5$ & $60.8$ & $85.4$ & $86.7$ & $57.3$ & $62.2$ & $47.2$ & $49.0$  \tabularnewline
% ~~~~-- LF \& Schema & $35.1$ & $37.4$ & $82.5$ & $84.6$ & $54.4$ & $61.2$ & \ \ $6.6$ & \ \ $7.0$ \tabularnewline
% ~~~~-- LF \& Schema \& DC & $30.9$ & $32.4$ & $80.5$ & $82.3$ & $48.7$ & $53.6$ & \ \ $1.8$ & \ \ $1.9$ \tabularnewline

% Exemplary Logical Forms & 65.3 & 70.0 & 73.1 & 76.3 & 54.0 & 57.7 & 66.6 & 72.3 \tabularnewline  % RnG logical forms
\bottomrule
\end{tabular}
\caption{Ablation experiments on the GrailQA validation set.
\textit{ELF} denotes retrieving exemplary logical forms as contexts.
\textit{Schema} denotes retrieving schema items as contexts.
\textit{CD} denotes constrained decoding in beam search.
\textit{Exemplary Logical Forms} means only using retrieved logical forms without generation.
}
\label{table:grailqa validation results}
\end{table*}

\subsection{Overall Performance}
\label{subsec:overall performance}

TIARA results on the GrailQA hidden test set and WebQSP test set are presented in Table \ref{table:grailqa test results} and Table \ref{table:webqsp test results}, respectively.

\paragraph{GrailQA Performance}
TIARA outperforms previous methods overall and in all three levels of generalization on both EM and F1.
Although previous methods \cite{chen21retrack,ye21rng,gu22dynamic} perform well in i.i.d., TIARA with multi-grained contexts gains an advantage over them in generalization scenarios.
% Though i.i.d. performance of previous methods is close to TIARA, TIARA performs better on generalization settings. 
It increases the F1 by 5.3 and 4.7 points compared to the previous SOTA in compositional and zero-shot generalization, respectively.
ArcaneQA \cite{gu22dynamic} is flexible in generating query structures with the help of dynamic program induction and has outperformed previous SOTA \cite{ye21rng} in compositional generalization by a large margin.
However, TIARA further boosts the performance in compositional generalization to 76.5 F1 points. % using the same PLM. (Generation model of Arcane is not PLM)
Compared to the improvement in F1 mentioned above, the improvement in EM is more significant, indicating that TIARA performs better in understanding questions, e.g., case \uppercase\expandafter{\romannumeral2} in Section \ref{subsec:case study}.
% The overall performance of TIARA on GrailQA validation set also outperforms previous methods, as shown in Table \ref{table:grailqa validation results}.
% It shows similar trends to those in the test set.

\paragraph{WebQSP Performance} 
TIARA outperforms existing IR- and SP-based methods in F1, without assumptions of a fixed number of hops nor using oracle entity annotations.
\citet{zhang22subgraph} propose sub-graph retrieval enhanced reasoning, which makes hits@1 substantially outperform other methods, but its F1 is still inferior to TIARA.
\citet{cao22program} transfer the knowledge from KQA Pro \cite{cao22kqa} to WebQSP with oracle entity annotations, which demonstrates the effect of pre-training on large KBQA datasets.
TIARA reaches comparable results without such extra refinement.
For further comparison, we also evaluate TIARA's performance given oracle annotations, which boosts the F1 score to 78.9 points.
% It still outperforms previous methods with oracle assumptions, which boosts the F1 score to $78.9$ points.
Besides the methods listed in Table \ref{table:webqsp test results}, Unik-QA \cite{oguz2020unik} can achieve 76.7\% Hits@1 on WebQSP by using T5-base to generate the answer text. However, while its results on WebQSP would be on par with TIARA, leveraging PLMs in an open-domain QA manner cannot generate all answers and does not fully solve KBQA, since the answer text may be ambiguous for entity linking (Freebase entity texts are not unique).

\section{Analysis}
\label{sec:analysis}

\subsection{Ablation Study}
\label{subsec:ablation study}

To verify the effect of the proposed multi-grained contextual retrieval and constrained decoding, we conduct ablation experiments using the GrailQA validation set, as shown in Table \ref{table:grailqa validation results}.
The ablation experiment to remove the exemplary logical form retrieval (w/o ELF) or schema retrieval (w/o Schema) is to retrain and inference with different contexts. 

\paragraph{Generation Alleviates Deficiencies of Retrieval}
% We evaluate the QA performance of retrieved exemplary logical forms without generation.
Using exemplary logical forms directly as prediction can achieve 72.9 F1 points, but they perform poorly in compositional generalization with complex structures (60.7 F1 points).
This issue is well mitigated by generation, which achieves 72.1 F1 points w/o Schema \& CD.
TIARA with multi-grained contexts further gains 2.7 F1 points in compositional setting.

\paragraph{Exemplary Logical Forms as Critical Contexts}
TIARA w/o ELF decreases performance by 26.5 F1 points in zero-shot setting, indicating the critical role of exemplary logical forms. % in the generation.
PLMs have the transferable understanding capability but cannot perceive KB structure, which can be largely alleviated by leveraging connected structures in exemplary logical forms.
% to the PLM is essential for generating high-quality logical forms.

\paragraph{Schema Retrieval as Necessary Supplement}
% However, logical form retrieval is not always available (Section \ref{subsec:multi-grained retriever}), and schema retrieval can provide an important supplement.
Removing schema retrieval (w/o Schema) decreases F1 by 2.7 points overall , and even 25.2 points (w/o ELF \& Schema compared to w/o ELF) when exemplary logical forms are not considered.
It shows schema items effectively provide semantics.
Besides, TIARA with no context other than entities (w/o ELF \& Schema \& CD) performs very poorly in zero-shot setting (F1 only 2.3 points), while schema retrieval remedies this issue to a large extent (50.6 F1 points w/o ELF \& CD).

\paragraph{Constrained Decoding Reduces Errors}
When the semantics and structure from exemplary logical forms and schema items are effective, constrained decoding contributes to 0.4 F1 points overall.
The PLM performance drops dramatically when these contexts are unavailable, and TIARA w/o ELF \& Schema improves F1 by 5.2 points compared to TIARA w/o ELF \& Schema \& CD.
Constrained decoding without any additional retrieval or training can reduce generation errors, especially when no context is available for PLM augmentation.

\paragraph{Mention Detection as Key to Entity Recall}
%\paragraph{Mention Detection boosts Model Recall}
\label{subsec:the impact of entity retriever}

Mention detection (i.e., the first step of entity retrieval) is the key to entity recall since it determines the upper-bound recall of the entity retriever. 
In this step, we adopt SpanMD as mention detector (described in Section \ref{subsec:multi-grained retriever}) instead of the typically used BERT-NER in previous works \cite{gu21beyond,chen21retrack,ye21rng,gu22dynamic}.
It shows that replacing BERT-NER with SpanMD leads to 2.2 F1 points improvement overall. 
More specifically, SpanMD obtains a notable gain in the zero-shot scenario (F1 +3.4 points), demonstrating the superiority of SpanMD in tackling zero-shot cases.
%Among the three settings, 
% SpanMD substantially outperforms BERT-NER by 2.2 points on recall overall and, more specifically, it achieves a significant recall gain (4.7 points) in the zero-shot setting compared to BERT-NER. As a consequence, our entity retriever significantly outperforms all the previous models by at least 5.0 F1 points on the GrailQA validation set. %the shorter mention from pairs of mentions that have the same candidates, which outperforms the strongest baseline \cite{ye21rng} by $1.43$ points%demonstrating its superiority in tackling zero-shot cases.
%
%Noticeably, although SpanNER causes the slightly inferior precision in the first step, the following step (i.e., candidate generation) boosts the overall precision from $79.2$ points to $88.8$ points via removing unreasonable mentions. Then after the entity disambiguation, our model significantly outperforms all the baseline models by at least $5$ F1 points on the GrailQA validation set. %the shorter mention from pairs of mentions that have the same candidates, which outperforms the strongest baseline \cite{ye21rng} by $1.43$ points.
%
%Overall, after the third step of entity linking, i.e., disambiguation, our model performs significantly better against all baselines and boosts \cite{ye21rng} by $5$ F1 points.
% Due to the limited space, please refer to Appendix \ref{sec:entity retrieval performance} for more discussion on entity retrieval.
%
Please refer to Appendix \ref{sec:entity retrieval performance} for more discussion on mention detection and entity retrieval.

\begin{table}[ht]
\centering
\small
\begin{tabular}{p{0.9cm}p{1.2cm}<{\centering}p{1.2cm}<{\centering}p{1.2cm}<{\centering}p{1.2cm}<{\centering}}
% \begin{tabular}{p{0.9cm}cccc}
\toprule
\textbf{Function} & \textbf{None} & \textbf{Count} & \textbf{Comp.} & \textbf{Super.} \\ \midrule
% ArcaneQA       & $70.8$/$77.8$ & $62.5$/$68.2$ & $54.5$/$75.7$ & $70.5$/$75.6$ \\
RnG       & 77.5/81.8 & 73.0/77.5 & 55.1/76.0 & 13.8/22.3 \\
\textbf{TIARA} & \textbf{77.8}/\textbf{83.1} & \textbf{76.4}/\textbf{81.8} & \textbf{57.4}/\textbf{81.4} & \textbf{58.7}/\textbf{69.0}\\ 
\midrule
\textbf{\#relation} & \textbf{1} & \textbf{2} & \textbf{3} & \textbf{4} \\ \midrule
RnG       & 75.7/79.3 & \textbf{65.3}/74.7 & 28.6/44.5 & \textbf{100.0}/\textbf{100.0} \\
\textbf{TIARA} & \textbf{81.2}/\textbf{85.6} & 64.7/\textbf{75.8} & \textbf{29.3}/\textbf{48.5} & 50.0/83.3 \\ 
\midrule
%\textbf{\#entity} & \textbf{0} & \textbf{1} & \textbf{2} & \\ \midrule 
%RnG-KBQA & $58.5$/$63.6$ & $75.4$/$79.9$ & $\mathbf{55.6}$/$\mathbf{73.5}$ & \\
%\textbf{TIARA}  & $\mathbf{77.5}$/$\mathbf{83.1}$  & $\mathbf{76.6}$/$\mathbf{82.6}$ & $49.9$/$68.0$ \\ 
%\bottomrule
\end{tabular}
\caption{EM and F1 results (\%) of different types of questions on the GrailQA validation set. \textit{RnG} denotes RnG-KBQA \cite{ye21rng}.
\textit{None} denotes only operators \texttt{AND} and \texttt{JOIN} are in the s-expression. 
\textit{Comp.} and \textit{Super.} denotes comparative and superlative. 
% \textit{\#relation} and \textit{\#entity} denote the number of relations and entities in the s-expression.
\textit{\#relation} denotes the number of relations in the s-expression.
}
\label{tab:detail analysis}
\end{table}

\begin{table*}[h]
\centering
\small
\begin{tabular}{p{0.97\textwidth}}
\toprule
\textbf{Case \uppercase\expandafter{\romannumeral1} Question} name the system that has decimetre as a measurement unit. \tabularnewline
\textbf{TIARA} (AND measurement\_unit.measurement\_system (JOIN measurement\_unit.measurement\_system.\textcolor{blue}{\textbf{length\_units}} m.01p5ld)) (\Checkmark) \tabularnewline
\textbf{TIARA w/o ELF} (AND measurement\_unit.measurement\_system (JOIN measurement\_unit.measurement\_system.
\tabularnewline
\textcolor{red}{\textbf{substance\_units}} m.01p5ld))
 (\XSolidBrush) \tabularnewline
\midrule
\textbf{Case \uppercase\expandafter{\romannumeral2} Question} which bipropellant rocket engine has a chamber pressure of less than 257.0 and uses an oxidizer of lox? \tabularnewline
\textbf{TIARA} (AND spaceflight.bipropellant\_rocket\_engine (AND \textcolor{blue}{\textbf{(JOIN spaceflight.bipropellant\_rocket\_engine.oxidizer m.01tm\_5)}}
(\textcolor{blue}{\textbf{lt}} spaceflight.bipropellant\_rocket\_engine.chamber\_pressure 257.0\^{}\^{}float))) (\Checkmark) 
\tabularnewline
\textbf{TIARA w/o Schema} (AND spaceflight.bipropellant\_rocket\_engine (\textcolor{red}{\textbf{JOIN}} spaceflight.bipropellant\_rocket\_engine. \tabularnewline
chamber\_pressure 257.0\^{}\^{}float)) (\XSolidBrush) \tabularnewline
\midrule
\textbf{Case \uppercase\expandafter{\romannumeral3} Question} find the smallest possible unit of resistivity. \tabularnewline
\textbf{TIARA} (ARGMIN measurement\_unit.unit\_of\_\textcolor{blue}{\textbf{resistivity}} measurement\_unit.unit\_of\_resistivity.resistivity\_in\_ohm \tabularnewline
\_meters) (\Checkmark) \tabularnewline
\textbf{TIARA w/o CD} (ARGMIN measurement\_unit.unit\_of\_\textcolor{red}{\textbf{resistance\_unit}} measurement\_unit.unit\_of\_resistivity.resistivity
\tabularnewline
\_in\_ohm\_meters (\XSolidBrush) \tabularnewline
\bottomrule
\end{tabular}
\caption{Case study of predicted logical forms by TIARA variants on the GrailQA validation set. 
The differences in the predictions are \textbf{bolded}. 
Errors are marked in \textcolor{red}{\textbf{red}}, and the corresponding correct parts are marked in \textcolor{blue}{\textbf{blue}}.}
\label{tab:case study}
\end{table*}

\subsection{In-Depth Analysis}
\label{subsec:in-depth analysis}
To show how TIARA performs on different logical form functions and structures, we compare it with previous SOTA \cite{ye21rng} on the GrailQA validation set in more detail, as shown in Table \ref{tab:detail analysis}.

\paragraph{Function} 
TIARA performs better for all function types even with limited data, especially for superlative questions. 
While some superlative questions have no entities, which leads to no exemplary logical forms, TIARA does not rely exclusively on these logical forms and still has schema items as contexts. 
Thus, it reaches a 3x improvement on F1 score for this function type.

\paragraph{\#relation} 
TIARA also performs well when considering different numbers of relations, except only for the number of relations being 4 (due only to two questions, one of which is incorrect because of a missing numerical property).

% \paragraph{\#entity.}
% TIARA has a significant advantage when \#entity is 0 (F1 $+19.5\%$), and the reason is similar to that described above.
% However, when generating a logical form containing two entities, it may be disturbed by schema retrieval, which degrades the performance. 
% As a comparison, in exemplary logical forms of RnG-KBQA, two entities should be connected if they both exist, making the model tend to generate executable results.

\subsection{Error Analysis}
\label{subsec:error analysis}

To further explore the limitations of TIARA, we randomly sample 50 questions whose predictions are different from the ground truth in the GrailQA validation set. 
The major errors can be classified as follows: \textbf{Entity retriever errors (46\%)} is caused by mention detection failure or high ambiguity of mentions, e.g., TIARA misses ``volt per metre'' in the question ``what is the unit of volt per metre?''. 
Entity retrieval remains a key challenge for KBQA.
\textbf{Syntactic errors (26\%)} mostly occurs when questions involve 1) rare operators, e.g., \texttt{ARGMIN}, and \texttt{ARGMAX} (5.95\% in the training corpus);
2) multiple constraints, whose logical forms are relatively complicated.
%3) unseen schema items, never appear in the training corpus. %it is challenging for the PLM when provided with several semantically similar zero-shot schema items %for example, ``which locomotive class is built by alstom coradia juniper?'' implicitly refers to ``rail.locomotive\_class.parent\_class''.
%However, TIARA generates ``rail.locomotive\_class.subclasses''.
%For example, the topic entity in the question ``what is the unit of volt per metre?''  is ``volt per metre'', while TIARA misses the correct mention. 
%Of entity retriever errors, $73.9\%$ are from zero-shot questions, and entity retriever drops around $10$ points F1 from i.i.d. to zero-shot cases. 
%\textbf{Syntactic error (26\%)} almost appears in the questions involving: 
%1) superlative operations (e.g., \texttt{ARGMIN}, \texttt{ARGMAX}), which only accounts for 5.95\% of the training data; 
%2) multiple constraints, whose logical forms are relatively complicated, e.g., ``the leader of both greek orthodoxy and the religion that worships church of notre-dame-des-arts is who?''.
\textbf{Semantic errors (12\%)} where TIARA selects incorrect schema items when correct ones are provided. Of those errors of this type, 83.3\%  come from zero-shot instances.
%For example, ``which locomotive class is built by alstom coradia juniper?'' implicitly refers to ``rail.locomotive\_class.parent\_class''.
%However, TIARA generates ``rail.locomotive\_class. subclasses''. 
%Of this type, 
\textbf{False negatives (6\%)} mainly occurs in comparative questions, i.e., ``larger than'' is annotated as $\geq$ in some cases.
\textbf{Miscellaneous (10\%)} occurs when the question is semantically ambiguous, and predicted results are inconsistent with the oracle schema in the KB. % understanding results
%For example, "which radio programs focus on music?" is linked to ``broadcast.genre'' in the KB.
%However, TIARA returns ``radio.radio\_program'' whose semantics is closer to the question.

%For example, the ground truth is \textit{greater than and equal to} ($\geq$) for ``what tropical cyclone has a minimum speed larger than 225.0?''.
%However, TIARA predicts \textit{greater than} ($>$). 

%The bottleneck of our model is the entity linking, especially for zero-shot cases.  

\subsection{Case Study}
\label{subsec:case study}

To aid in visualizing TIARA's capabilities, we select three case studies from the GrailQA validation set in Table \ref{tab:case study}.
In Case \uppercase\expandafter{\romannumeral1}, the entity m.01p5ld (decimetre) is not connected to the relation ``substance\_units'' predicted by \textbf{TIARA w/o ELF}.
%, and the correct relation ``length\_units'' is not in the results of schema retrieval.
Using only discrete schema does not guarantee that generated logical forms are connected.
The exemplary logical forms (ELF) ensure graph connectivity and provide better contexts in this case.
Case \uppercase\expandafter{\romannumeral2} contains an entity and a literal connected by two relations, as well as a function \textit{lt} ($<$). 
\textbf{TIARA w/o the Schema} chooses one of its exemplary logical forms with only one relation and no function.
% the logical form generated by \textbf{TIARA w/o Schema} is only limited to two hops, while the correct logical form has three clauses has one more clause than it does.
The enumeration of exemplary logic forms is limited regarding literals and functions, but schema retrieval contributes to a semantic supplement.
Lastly, in Case \uppercase\expandafter{\romannumeral3}, \textbf{TIARA w/o CD} selects the phrase ``resistance\_unit'' that, while somewhat semantically similar to the correct one, does not exist in the KB. % when predicting the schema item.
It shows that constrained decoding helps PLM to generate logical forms that conform to the KB specification. % semantics of logical forms.

% \subsection{Further Exploration}
% \label{subsec:future exploration}

% When TIARA's T5-base is replaced with T5-large \cite{raffel20exploring} while keeping the other settings unchanged, its F1 score on the GrailQA validation set improves from $81.9\%$ to $83.0\%$. 
% PLMs with a larger number of parameters should contribute more to the KBQA performance and deserve further discussion.

% Recall the pilot experiment mentioned in Section \ref{sec:introduction}, TIARA achieves up to $92\%$ F1 on GrailQA validation set given oracle entities, schema items, and exemplary logical forms.
% TIARA has great potential if the retrieval performance is further improved.

\section{Conclusion}
\label{sec:conclusion}

We present TIARA to empower PLMs on QA over large-scale KBs.
Multi-grained retrieval transcends insufficient exemplary logical forms and isolated schema items, and its results provide both semantic and syntactic contexts for a given question. 

In systematic stage-by-stage analysis, we show that: 
i) constrained decoding prunes invalid search and reduces generation errors (e.g., non-existent schema and illegal operators); 
ii) entity linking improvements address a key bottleneck for zero-shot generalization; 
iii) schema retrieval with flexible scope is an essential semantic supplement to surpass constraints of previous systems\footnote{E.g., limited to 2-hop relations; unable to deal with questions without entities.}; 
and iv)
%We demonstrate that comprehensive KB context requires multi-grained retrieval for coverage. 
the generalization capabilities of PLM can be further empowered by both the retrieval and constrained decoding for semantic parsing pipelines.
All of which combined result in an effective and robust KBQA system.
% for question answering over knowledge base.
% TIARA provides an idea to better exploit the potential of PLMs to parse the natural language question to its corresponding logical form. 
% Retrieving multiple types of KB context associated with a question can significantly augment the generation.
% Constrained decoding is also effective in reducing invalid generation.

Though we have started exploiting the potential of PLMs, there is still a gap between the pre-training of PLMs on natural language and the downstream semantic parsing task. 
In the future, it is worth exploring how to build a bridge between unstructured natural language and structured KB in the pre-training phase.
% In the future, it is worth exploring how to pre-train for KBQA tasks to make PLM adaptable to logical form generation tasks.
% it is worth exploring how to pre-train the KB for the semantic parsing task.

% In our pilot experiments, we find a gap between given the oracle KB context and PLM generation results. We suspect that the gap comes from the unstructured pre-training corpus and downstream structured data. We leave the pretraining on structured data as the future work.

\section*{Limitations}
\label{sec:limitations}

Our method learns from logical form annotations, which require expensive and specialized crowdsourcing work in data collection.

The retrieval efficiency of our method also deserves further optimization. 
Using figures from the evaluations over GrailQA, logical form enumeration alone takes more than 7 seconds per question when there is no cache, which would need improvement to meet requirements in certain practical scenarios.
Schema retrieval takes 1.41 seconds per question, which is the major additional time consumption compared to RnG-KBQA.
For comparison, BERT + \Ranking \cite{gu21beyond}, RnG-KBQA \cite{ye21rng}, and ArcaneQA \cite{gu22dynamic} take on average 115.5, 82.1 and 5.6 seconds per question respectively - as reported by \citet{gu22dynamic}.
ReTraCk \cite{chen21retrack} reports taking on average 1.62 seconds per question.

Lastly, while the semantic parsing approach in this paper applies well to RDF graphs, it is not fully applicable to other structured data, such as tables or unstructured data, which are important sources of knowledge in the wild.

\section*{Acknowledgments}
We would like to thank all the anonymous reviewers for their constructive comments and useful suggestions.
We thank Yu Gu for evaluating our submissions on the test set of the GrailQA benchmark.
We also thank Yu Gu and Xi Ye for sharing pre-processed data on GrailQA and WebQuestionsSP.

\bibliography{anthology,custom}
\bibliographystyle{acl_natbib}

\newpage
\clearpage

\appendix
\section{Retrieval Performance}
\label{sec:retrieval performance}

\subsection{Entity Retrieval Performance}
\label{sec:entity retrieval performance}

\begin{table}[hp]
\centering
\small
\begin{tabular}{lcccc}
\toprule
\textbf{Method} & \textbf{Overall} & \textbf{I.I.D.} & \textbf{Comp.} & \textbf{Zero.}\tabularnewline \midrule
BERT-NER & 92.1 & 98.2 & \textbf{96.9} & 87.5 \tabularnewline
\textbf{SpanMD} (Ours) & \textbf{94.3} & \textbf{98.3} & 95.1 & \textbf{92.2} \tabularnewline
\bottomrule
\end{tabular}
\caption{Recall (\%) of mention detection on the GrailQA validation set. \textit{Comp.} denotes compositional. 
\textit{Zero.} denotes zero-shot.
}
\label{table:MD recall performance}
\end{table}

\begin{table}[hp]
\centering %
\small
\begin{tabular}{lccc}
\toprule 
\textbf{Method} & \textbf{P} &  \textbf{R} & \textbf{F1} \tabularnewline \midrule 
% \multicolumn{4}{c}{\textit{GrailQA}}\tabularnewline
GrailQA~\cite{gu21beyond}      & 76.0 & 81.1 & 73.1 \tabularnewline
% ReTraCk~\cite{chen21retrack}   & 78.9 & 76.2 & 77.5 \tabularnewline
RnG-KBQA~\cite{ye21rng}        & 84.1 & 86.8 & 80.4 \tabularnewline
\textbf{TIARA} (Ours)           & \textbf{87.2} & \textbf{88.6} & \textbf{85.4} \tabularnewline
% \textbf{\multicolumn{4}{c}{\textit{WebQSP}} \tabularnewline
% ELQ~\citep{li20efficient} & $94.4$ & $89.7$ & $90.8$ \tabularnewline}
\bottomrule 
\end{tabular}
\caption{Precision (P), recall (R) and F1 (\%) of entity retrieval (\%) on the GrailQA validation set.}
\label{table:entity linking performance}
\end{table}

We report the recall of mention detection on the GrailQA validation set based on two different mention detectors to show the effectiveness of SpanMD. As shown in Table \ref{table:MD recall performance}, SpanMD substantially outperforms BERT-NER by 2.2 points on overall recall.\footnote{ \url{https://github.com/kamalkraj/BERT-NER}.} Specifically, SpanMD achieves a significant recall gain (4.7 points) in a more challenging zero-shot setting compared to BERT-NER, demonstrating its superiority in tackling zero-shot cases. 

In addition, we evaluate the overall performance of our entity retriever on the GrailQA validation set and report it in Table \ref{table:entity linking performance}. Specifically, we compare against the following baselines: 1) \textbf{GrailQA} \cite{gu21beyond} which simply chooses the most popular entity according to the prior features provided by FACC1 \cite{gabrilovich13facc1}. 
%
%2) \textbf{ReTrack} \cite{chen21retrack} which leverages \BOOTLEG~\citep{orr2020bootleg} to enrich the prior features and cope with long-tail entities. 
%
2) \textbf{RnG-KBQA} \cite{ye21rng} which utilizes the relation information linked with an entity to further enhance the semantics of the entity. While those and many previous works derive better entity retrieval results via focusing on how to improve the performance of disambiguation, TIARA also adopts an improved mention detector to boost entity retrieval. 

From Table \ref{table:entity linking performance}, it can be observed that our entity retriever significantly surpasses all the baseline models by at least 5.0 F1 points on the GrailQA validation set. Compared to the strongest baseline \cite{ye21rng}, we yield better results on both precision and recall, further demonstrating the effectiveness of the entity retriever.

Analogous to other previous SOTA \cite{ye21rng}, we use the off-the-shelf entity linker ELQ \cite{li20efficient} on WebQSP. ELQ is an end-to-end entity linking system that performs mention detection and entity disambiguation jointly on questions in one pass of BERT \cite{devlin19bert}. It is trained on both Wikipedia and WebQSP data and obtains SOTA performance on WebQSP. We refer readers to \citet{li20efficient} for more details.

\subsection{Schema Retrieval Performance}
\label{sec:schema retrieval performance}

\begin{table}[h]
\centering
\small
\begin{tabular}{lcccc}
\toprule
\textbf{Class} & \textbf{Overall} & \textbf{I.I.D.} & \textbf{Comp.} & \textbf{Zero.}\tabularnewline \midrule
ReTraCk & 94.3 & 98.1 & 97.5 & 91.3 \tabularnewline
\textbf{TIARA} & \textbf{95.8} & \textbf{99.6} & \textbf{97.9} & \textbf{93.4} \tabularnewline
\midrule
\textbf{Relation} & \textbf{Overall} & \textbf{I.I.D.} & \textbf{Comp.} & \textbf{Zero.} \tabularnewline \midrule
ReTraCk & 88.4 & 95.3 & 91.0 & 84.3 \tabularnewline
\textbf{TIARA} &  \textbf{92.0} & \textbf{97.9} & \textbf{93.7} & \textbf{88.7} \tabularnewline
\bottomrule
\end{tabular}
\caption{Recall (\%) of top-10 schema retrieval on the GrailQA validation set.}
\label{table:schema retrieval performance}
\end{table}

We report the schema retrieval performance for GrailQA in Table \ref{table:schema retrieval performance}.
TIARA's top-10 schema recall outperforms ReTraCk \cite{chen21retrack}, the highest-ranked system in the GrailQA leaderboard to apply a dense retriever for schema retrieval, across all three levels of generalization.
Note that ReTraCk uses 100 classes and 150 relations for each question, which does not result in higher QA performance, as shown in Table \ref{table:grailqa test results} and Table \ref{table:webqsp test results}.

\section{Performance on the WebQSP Validation Set}
\label{sec:performance on the webqsp validation set}

\begin{table}[ht]
\centering %
\small
\begin{tabular}{lc@{\hspace{3pt}}c}
\toprule 
\textbf{Method} & \textbf{F1} &  \textbf{Hits@1}\tabularnewline
\midrule 
\textbf{TIARA} & \textbf{76.7} & \textbf{74.5}  \tabularnewline
~~~~w/o Schema & 75.4 & 73.5 \tabularnewline
~~~~w/o ELF & 65.3 & 64.0 \tabularnewline
~~~~w/o ELF \& Schema & 63.9 & 61.5 \tabularnewline
% \midrule
% \textbf{TIARA}$^*$ & \textbf{76.2} & \textbf{74.0}  \tabularnewline
% ~~~~w/o Schema & 75.4 & 73.5 \tabularnewline
% ~~~~w/o ELF & 69.3 & 66.5 \tabularnewline
% ~~~~w/o ELF \& Schema & 67.0 & 63.5 \tabularnewline
\bottomrule 
\end{tabular}
\caption{F1 and hits@1 results (\%) on the WebQSP validation set.
}
\label{table:webqsp validation results}
\end{table}

We report the performance on the WebQSP validation set, as shown in Table \ref{table:webqsp validation results}. 
WebQSP does not contain the official validation set, and the dataset splitting is described in Section \ref{subsec:setup}.

\section{Settings and Hyperparameters}
\label{sec:parameters and hyperparameters}

Experimental settings are defined along the lines of previous works \cite{gu21beyond,chen21retrack,ye21rng} and empirically tuned non-exhaustively as significant results are obtained.
We manually tune the hyperparameters based on the validation set results.
The criterion used to select among them is F1 score.
Specifically, we search for the batch size and learning rate of the generator from [2, 4, 6, 8] and [1e-5, 3e-5, 5e-5].

The number of parameters in our models is 110M for BERT-base-uncased \cite{devlin19bert} and 220M for T5-base \cite{raffel20exploring}.

\section{Evaluation Details}
\label{sec:evaluation details}

We use official scripts to evaluate on both datasets\footnote{GrailQA script: \url{https://worksheets.codalab.org/bundles/0x2d13989c17e44690ab62cc4edc0b900d/}, and WebQSP script: \url{https://www.microsoft.com/en-us/download/details.aspx?id=52763}.}, except for hits@1 on WebQSP, as mentioned in Section \ref{subsec:setup}.

\end{document}